\documentclass[letterpaper, 10 pt, conference]{ieeeconf}  

\IEEEoverridecommandlockouts                              

\usepackage{amsmath} 
\usepackage{amssymb}  

\usepackage{cite}
\usepackage{acro}
\usepackage{siunitx}
\usepackage{bm}
\usepackage{nicefrac}
\usepackage{algorithm2e}
\RestyleAlgo{ruled}
\usepackage{bbm}
\usepackage{pict2e}
\usepackage{pgfplots}
\usepackage{multirow}
\usepackage{comment}
\usepackage{multicol}
\usepackage{tabularx}
\usepackage{svg}

\usepackage{dsfont}

\usepackage{makecell}

\usepackage{graphicx}
\usepackage{subcaption}
\usepackage[font=footnotesize]{caption}  
\usepackage[caption=false,font=footnotesize]{subfig}

\usepackage{svg}

\usepackage{placeins}
\pgfplotsset{compat=newest}

\acsetup{patch/maketitle=false}
\DeclareAcronym{AGV}{
    short = AGV,
    long = Autonomous Ground Vehicle
}

\DeclareAcronym{DRL}{
    short = DRL,
    long = Deep Reinforcement Learning
}
\DeclareAcronym{SFM}{
    short = SFM,
    long = Social Force Model
}

\DeclareAcronym{ORCA}{
    short = ORCA,
    long = Optimal Reciprocal Collision Avoidance
}

\DeclareAcronym{POMDP}{
    short = POMDP,
    long = Partially Observable Markov Decision Process
}

\DeclareAcronym{LSTM}{
    short = LSTM,
    long = Long Short-Term Memory
}

\DeclareAcronym{PPO}{
    short = PPO,
    long = Proximal Policy Optimization
}

\DeclareAcronym{ODV-PPO}{
    short = ODV-PPO,
    long = Observation-Dependent Variance PPO
}

\DeclareAcronym{SAC}{
    short = SAC,
    long = Soft Actor-Critic
}

\DeclareAcronym{NN}{
    short = NN,
    long = Neural Network
}

\DeclareAcronym{OOD}{
    short = OOD,
    long = Out-Of-Distribution
}

\DeclareAcronym{MC-Dropout}{
    short = MC-dropout,
    long = Monte-Carlo dropout
}

\DeclareAcronym{MPC}{
    short = MPC,
    long = Model-Predictive Controller

}
\DeclareAcronym{ODV}{
    short = ODV,
    long = Observation-Dependent Variance
}
\DeclareAcronym{MSE}{
    short = MSE,
    long = Mean Squared Error
}
\DeclareAcronym{DQN}{
    short = DQN,
    long = Double Q-Network
}

\DeclareAcronym{PCN}{
    short= PCN,
    long = Probability of Collision Network,
}

\DeclareAcronym{POC}{
    short= POC,
    long = Probability of Collision,
}

\DeclareAcronym{SCA}{
    short= SCA,
    long = Social Collision Avoidance,
}

\DeclareAcronym{SA}{
    short= SA,
    long = Socially Aware
}

\DeclareAcronym{SI}{
    short= SI,
    long = Socially Integrated,
}

\begin{document}

\title{
\vspace{-3.5em}\footnotesize This work has been accepted for publication at the \\
IEEE/RSJ International Conference on Intelligent Robots and Systems (IROS), 2025, \textcopyright IEEE \\\vspace{4.25em}
\LARGE \bf
Disentangling Uncertainty for Safe Social Navigation using Deep Reinforcement Learning
}

\author{Daniel Flögel$^{*1}$, Marcos Gómez Villafañe$^{*1,2}$, Joshua Ransiek$^{1}$, and Sören Hohmann$^{3}$
\thanks{$^*$\,Authors contributed equally: Daniel Flögel,  Marcos Gómez Villafañe}
\thanks{$^{1}$ are with FZI Research Center for Information Technology, Karlsruhe, Germany
        {\tt\small floegel@fzi.de}}%
\thanks{$^{2}$ is with Facultad de Ingeninería, Universidad de Buenos Aires, Buenos Aires, Argentina}%
\thanks{$^{3}$ is with the Institute of Control Systems at Karlsruhe Institute of Technology, Karlsruhe, Germany
        {\tt\small soeren.hohmann@kit.edu}}%
}

\maketitle
\thispagestyle{empty}
\pagestyle{empty}

\begin{abstract}
Autonomous mobile robots are increasingly used in pedestrian-rich environments where safe navigation and appropriate human interaction are crucial.
While \ac{DRL} enables socially integrated robot behavior, challenges persist in novel or perturbed scenarios to indicate \textit{when and why} the policy is uncertain.
Unknown uncertainty in decision-making can lead to collisions or human discomfort and is one reason why safe and risk-aware navigation is still an open problem.
This work introduces a novel approach that integrates \textit{aleatoric}, \textit{epistemic}, and \textit{predictive} uncertainty estimation into a \ac{DRL} navigation framework for policy distribution uncertainty estimates.
We, therefore, incorporate \ac{ODV} and dropout into the \ac{PPO} algorithm.
For different types of perturbations, we compare the ability of deep ensembles and \ac{MC-Dropout} to estimate the uncertainties of the policy. 
In uncertain decision-making situations, we propose to change the robot's social behavior to conservative collision avoidance.
The results show improved training performance with \ac{ODV} and dropout in \ac{PPO} and reveal that the training scenario has an impact on the generalization. 
In addition, \ac{MC-Dropout} is more sensitive to perturbations and correlates the uncertainty type to the perturbation better.
With the safe action selection, the robot can navigate in perturbed environments with fewer collisions. 

\end{abstract}

\section{INTRODUCTION}
Autonomous mobile robots are increasingly being deployed in a variety of public pedestrian-rich environments, e.g. pedestrian zones for transport or cleaning 
\cite{AlaoUncertaintyawareNavigationinCrowded2022, SunRiskAwareDeepReinforcementLearningf2023 ,MavrogiannisCoreChallengesofSocialRobotNavigati2023}, while \ac{DRL}-based approaches are increasingly used \cite{MavrogiannisCoreChallengesofSocialRobotNavigati2023, GolchoubianUncertaintyAwareDRLforAutonomousVeh2024}.
These robots not only have socially aware behavior but also become socially integrated to reduce the negative impact on humans \cite{FloegelSociallyIntegratedNavigation2024}.
However, there are two fundamental challenges.
The first arises through the variety and a priori unknown scenarios caused by the stochastic behavior of humans \cite{ZhuDeepreinforcementlearningbasedmobile2021, RyuIntegratingPredictiveMotionUncertaint2024}.
The second through the general problem of machine learning-based systems to predict correct actions in unseen scenarios \cite{SedlmeierUncertaintyBasedOutofDistributionDe2020, FanLearningResilientBehaviorsforNavigat2020}. 
Since the robot has to choose an action in each step, one risk is a high level of uncertainty in the decision-making process at action selection \cite{EverettCertifiableRobustnesstoAdversarialSt2022}. 
As a consequence, safe and risk-aware social navigation to avoid collisions and human discomfort is still an open problem \cite{YangRMRLRobotNavigationinCrowdEnvironm2023, MavrogiannisCoreChallengesofSocialRobotNavigati2023}.

\begin{figure}[!htb]
    \vspace{0.4em}
    \centering
     \includegraphics[width=0.95\linewidth]{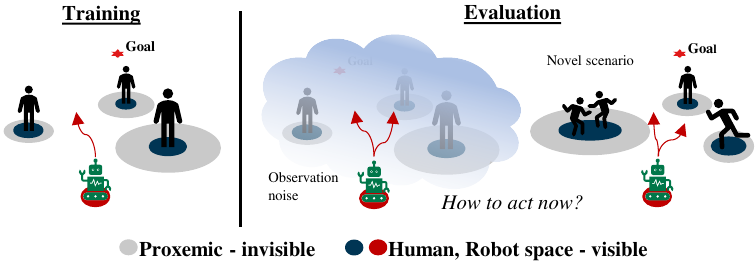}
    \vspace{-0.3em}
    \caption{We propose to integrate aleatoric and epistemic uncertainty estimation of action selection into \ac{DRL}-based navigation policies to detect scenarios where the robot is uncertain, e.g. due to observation noise or novel scenarios. Depending on the confidence about action selection, the robot either interacts with humans or focuses on safe collision avoidance. }
    \label{fig:paper_figure}
    \vspace{-1.2em}
\end{figure}

The uncertainty in the robot's decision-making is affected by scenarios the robot has never experienced in training, \ac{OOD} scenarios, and due to the inherent uncertainty of the environment, like sensor noise or occlusion \cite{MunOcclusionAwareCrowdNavigationUsingP2023}.
Thus, the \ac{DRL} policy will face \textit{epistemic uncertainty} (model uncertainty), which correlates with \ac{OOD}, and \textit{aleatoric uncertainty}, which stems from stochastic environments, e.g. sensor noise or perturbations on selected actions \cite{ClementsEstimatingRiskandUncertaintyinDeep2019}, as depicted in Fig~\ref{fig:paper_figure}.
For a \ac{DRL} policy, it is essential to estimate both types as it is expected that the robot identifies anomalous environmental states if it does not know which action to take to account for potential risks \cite{ClementsEstimatingRiskandUncertaintyinDeep2019, charpentier_disentangling}.
Thus, disentangling aleatoric and epistemic uncertainty is crucial for a \ac{DRL} policy towards safe and risk-aware behavior \cite{charpentier_disentangling}. 
However, there is less literature on uncertainty-aware \ac{DRL}, compared to supervised learning \cite{LockwoodAReviewofUncertaintyforDeepReinfor2022}, and only a few works address the resulting risks in social navigation \cite{ChoiRiskConditionedDistributionalSoftAct2021}.

Most \ac{DRL}-based navigation approaches, e.g. \cite{ZhuCollisionAvoidanceAmongDenseHeteroge2023 ,NarayananEWareNet:EmotionAwarePedestrianInten2023 ,FloegelSociallyIntegratedNavigation2024, ZhangRobotNavigationamongExternalAutonomo2020, SamsaniSociallyCompliantRobotNavigationinC2021, EverettCollisionAvoidanceinPedestrianRichE2021},
do not consider \ac{OOD} scenarios or noise, which is a significant risk since the \ac{OOD} performance of recent approaches is poor as evaluated in \cite{MavrogiannisCoreChallengesofSocialRobotNavigati2023}.
For known input perturbations, a certified adversarial robustness analysis framework for \ac{DRL} is proposed in \cite{EverettCertifiableRobustnesstoAdversarialSt2022}.
However, the perturbation must be known a priori, and it only applies to \ac{DRL} approaches with discrete actions.
Other recent works make use of the uncertainty in the human trajectory prediction \cite{GolchoubianUncertaintyAwareDRLforAutonomousVeh2024, AlaoUncertaintyawareNavigationinCrowded2022} or propose to detect novel scenarios \cite{FanLearningResilientBehaviorsforNavigat2020}. 
However, the policy can generalize to other scenarios, and these approaches neglect the uncertainty in decision-making, which is the crucial part \cite{EverettCertifiableRobustnesstoAdversarialSt2022}.
In addition, \cite{FanLearningResilientBehaviorsforNavigat2020} intentionally replaced \ac{PPO} \cite{SchulmanProximalPolicyOptimizationAlgorithms2017} with \ac{SAC} \cite{HaarnojaSoftActorCriticAlgorithmsandApplica2018} because it is challenging to train \ac{PPO} with state-dependent variance of policy distribution while having a good exploration-exploitation trade-off.
However, many social navigation approaches rely on the \ac{PPO} algorithm due to the stable training and good performance, e.g.\cite{MunOcclusionAwareCrowdNavigationUsingP2023, FloegelSociallyIntegratedNavigation2024, SunRiskAwareDeepReinforcementLearningf2023, GolchoubianUncertaintyAwareDRLforAutonomousVeh2024, YaoCrowdAwareRobotNavigationforPedestr2021, BritoWheretogonext:LearningaSubgoalRec2021, HuSafeNavigationWithHumanInstructions2019, ChenSociallyAwareObjectGoalNavigationWi2024, LiMultiAgentDynamicRelationalReasoning2024, XieDRLVO:LearningtoNavigateThroughCro2023}.

We see the clear limitation that the uncertainty in the policy distribution is neither considered, nor is the source of the uncertainty determined.
Thus, we propose to disentangle the policy's uncertainty measures and to adapt the robot's interaction behavior according to the confidence in decision-making, as it is essential for successful deployment in real-world environments \cite{LockwoodAReviewofUncertaintyforDeepReinfor2022, ClementsEstimatingRiskandUncertaintyinDeep2019}.

As main contributions of this work, (i) we extend the \ac{PPO} algorithm with \acf{ODV} and ensure stable training with an additional loss function and variance clamping. 
(ii) On top of that, we compare the capability of \ac{MC-Dropout} and deep ensembles to disentangle the uncertainty estimates of the policy distribution and incorporate predictive uncertainty estimation into the feature extractor. 
(iii) Based on the uncertainty estimates, we use a parametrized function to detect high-risk scenarios and change the robot's social interaction behavior to reduce collisions.
We conduct the experiments in simulation to evaluate the algorithms' response to dynamic obstacles in a reproducible environment and investigate the impact of training scenarios on generalization, as we consider this as an essential prior step for real-world experiments.

\section{Related Work}
\subsection{Uncertainty Estimation in Machine Learning}

Estimating the uncertainty can be distinguished into sample-free and sample-based methods, e.g., \ac{MC-Dropout} or ensembles, which require multiple forward passes \cite{Charpentier_PHD}.
Ensemble methods consist of multiple models trained with different weight initializations.
The variance of the different model predictions can be considered as Bayesian uncertainty estimates \cite{Charpentier_PHD}. 
\ac{MC-Dropout}, which uses dropout during testing as regularization, randomly drops neurons in testing to form different models and combine their predictions as Bayesian uncertainty estimation \cite{GalDropoutasaBayesianApproximation:Rep2015}.
The uncertainty estimates can be further distinguished into aleatoric, epistemic, and predictive uncertainty \cite{GalPhD, LockwoodAReviewofUncertaintyforDeepReinfor2022}.
Aleatoric uncertainty arises from the stochastic nature of the environment and can not be reduced. 
The three primary sources in \ac{DRL} are stochasticity in observation, actions, and rewards \cite{LockwoodAReviewofUncertaintyforDeepReinfor2022}.
Epistemic uncertainty arises from limited knowledge gathered in training and accounts for the lack of knowledge of a policy, but can be reduced with more training samples \cite{Charpentier_PHD,LockwoodAReviewofUncertaintyforDeepReinfor2022}.
Predictive uncertainty summarizes the effects of aleatoric and epistemic uncertainty \cite{GalPhD,MucsanyiBenchmarkingUncertaintyDisentanglement2024}.
The literature for uncertainty-aware \ac{DRL} includes Bayesian \ac{DRL} \cite{GhavamzadehBayesianReinforcementLearningASurve2015}, such as an ensemble \cite{B.LutjensSafeReinforcementLearningWithModelU2019} or dropout \cite{KahnUncertaintyAwareReinforcementLearning2017} approaches.
However, only a few methods aim to disentangle the uncertainty estimation in action selection \cite{LockwoodAReviewofUncertaintyforDeepReinfor2022}. 
For the \ac{DQN} algorithm, aleatoric and epistemic uncertainty are disentangled in \cite{charpentier_disentangling, ClementsEstimatingRiskandUncertaintyinDeep2019} but not adapted for actor-critic algorithms with continuous action space such as \ac{PPO}, which is challenging \cite{FanLearningResilientBehaviorsforNavigat2020}.
The sample-based approximations, \ac{MC-Dropout} and ensembles, require the least changes to the \ac{PPO} algorithm, but challenges arise for aleatoric estimates as it requires an observation-dependent variance in policy distribution, which is not the case for the \ac{PPO} algorithm and not necessarily boosts training performance.
In \ac{DRL}, dropout is also not a common choice due to the non-stationary targets \cite{LockwoodAReviewofUncertaintyforDeepReinfor2022}, and naive application can be challenging for \ac{PPO} \cite{HausknechtConsistentDropoutforPolicyGradientR2022}.
In addition, a benchmark of different uncertainty estimation methods for a classification task shows that disentangling the uncertainties remains an unsolved problem. 
It concludes that tailored methods for specific tasks are needed \cite{MucsanyiBenchmarkingUncertaintyDisentanglement2024}. 
Thus, disentangling uncertainty estimates and incorporating them into \ac{DRL} is an ongoing field of research.

\subsection{Safety in Social Navigation}
\ac{DRL}-based social navigation approaches can be categorized based on the robot's exhibited social behavior in human-machine interaction into \ac{SCA} with a lack of social aspects, \ac{SA} approaches with a predefined social behavior, and \ac{SI} where the robot's behavior is adaptive to human behavior and emerges through interaction \cite{FloegelSociallyIntegratedNavigation2024}. 
For a more detailed description, the reader is referred to \cite{MavrogiannisCoreChallengesofSocialRobotNavigati2023}, \cite{RiosMartinezFromProxemicsTheorytoSociallyAware2015}, \cite{FloegelSociallyIntegratedNavigation2024} for core challenges, proxemic theory, and distinction of social navigation approaches, respectively. 

For safety in social navigation, early works proposed a \ac{POC} distribution in combination with a model-predictive controller (MPC) for safe and uncertainty-aware action selection \cite{B.LutjensSafeReinforcementLearningWithModelU2019, KahnUncertaintyAwareReinforcementLearning2017}.
\cite{KahnUncertaintyAwareReinforcementLearning2017} uses \ac{MC-Dropout} and bootstrapping to estimate the \ac{POC} distribution for a risk-aware exploration in model-based \ac{DRL}.
An ensemble of LSTMs is used in \cite{B.LutjensSafeReinforcementLearningWithModelU2019} in combination with \ac{MC-Dropout} and bootstrapping to estimate the distribution. 
A risk function is proposed in \cite{SunRiskAwareDeepReinforcementLearningf2023} to capture the \ac{POC} to prioritize humans with a higher risk of collision.
However, a learned \ac{POC} can be uncertain and does not reflect the policy uncertainty in the action selection. 
Other approaches use the uncertainty in human trajectory prediction for risk-aware planning \cite{GolchoubianUncertaintyAwareDRLforAutonomousVeh2024, AlaoUncertaintyawareNavigationinCrowded2022}.
Such approaches are highly susceptible to the stochasticity of noise and the unobserved intentions of the external agents, which is addressed in \cite{Umbrella} with a model-based \ac{DRL} approach by estimating the aleatoric uncertainty of the trajectory prediction.
A risk-map-based approach with human position prediction and probabilistic risk areas instead of hard collision avoidance is proposed in \cite{YangRMRLRobotNavigationinCrowdEnvironm2023} to address dynamic human behavior and static clutter. 
Other works propose safety zones around humans, e.g. \cite{KastnerEnhancingNavigationalSafetyinCrowded2021, SamsaniSociallyCompliantRobotNavigationinC2021}, to increase the minimum distance between the robot and humans. 
To overcome the problem of occluded humans, the social inference mechanism with a variational autoencoder to encode human interactions is incorporated in \cite{MunOcclusionAwareCrowdNavigationUsingP2023}.
A risk-conditioned distributional \ac{SAC} algorithm that learns multiple policies concurrently is proposed in \cite{ChoiRiskConditionedDistributionalSoftAct2021}. 
The distributional \ac{DRL} learns the distribution over the return and not only the expected mean, and the risk measure is a mapping from the return distribution to a scalar value.
Other work estimates uncertainty from environmental novelty \cite{RichterSafeVisualNavigationviaDeepLearning2017}, which does not translate to policy uncertainty.
A resilient robot behavior for navigation in unseen, uncertain environments with collision avoidance is addressed in \cite{FanLearningResilientBehaviorsforNavigat2020}.
An uncertainty-aware predictor for environmental uncertainty is proposed to learn an uncertainty-aware navigation network in prior unknown environments.

In summary, no existing model-free \ac{DRL} approaches disentangle and consider the policy distribution uncertainty in action selection, which is crucial for safe and risk-aware decision-making.

\section{Preliminaries}
A dynamic object in the environment is generally referred to as an agent, either a robot or a human, and a policy determines its behavior.
Variables referred to the robot are indexed with $x^0$, and humans with $x^i$ with $i \in 1, \cdots N-1$. 
A scalar value is denoted by $x$ and a vector by $\bm{x}$.

\subsection{Problem Formulation}
The navigation task of one robot toward a goal in an environment of $N-1$ humans is a sequential decision-making problem and can be modeled as \ac{POMDP} and solved with a \ac{DRL} framework \cite{ChenDecentralizednoncommunicatingmultiage2017}. 
The \ac{POMDP} is described with a 8-tuple $(\mathcal{S}, \mathcal{A}, \mathcal{T}, \mathcal{O}, \Omega, \mathcal{T}_0, R, \gamma)$.
We assume the state space $\mathcal{S}$ and action space $\mathcal{A}$ as continuous.
The transition function $\mathcal{T} : \mathcal{S} \times \mathcal{A} \times \mathcal{S} \rightarrow [0,1]$ describes the probability transitioning from state $\bm{s}_{t} \in \mathcal{S}$ to state $\bm{s}_{t+1} \in \mathcal{S}$ for the given action $\bm{a}_t \in \mathcal{A}$. 
With each transition, an observation $\bm{o}_t \in \mathcal{O}$ and a reward $R : \mathcal{S} \times \mathcal{A} \rightarrow \mathbb{R}$ are returned by the environment. 
The observation $\bm{o}_t$ is returned with probability $\Omega(\bm{o}_t|\bm{s}_t)$ depending on the sensors.
The initial state distribution is denoted by $\mathcal{T}_0$ while $\gamma \in [0,1)$ describes the discount factor.
Every agent is completely described with a state $\bm{s}_{t}^{i} = [\bm{s}_{t}^{i,\mathrm{o}} , \bm{s}_{t}^{i,\mathrm{h}}]$ at any given time $t$.
The state is separated into two parts. 
The observable part $\bm{s}_{t}^{i,\mathrm{o}} = [\bm{p}, \bm{v}, r]$ is composed of position $\bm{p} $, velocity $\bm{v}$, and radius $r$.
The unobservable, hidden part, $\bm{s}_{t}^{i,\mathrm{h}} = [\bm{p}_{\mathrm{g}}, v_{\mathrm{pref}}, \psi_{\mathrm{pref}}, r_{\mathrm{prox}}$] is composed of goal position $\bm{p}_{\mathrm{g}}$, preferred velocity $v_{\mathrm{pref}}$, preferred orientation $\psi_{\mathrm{pref}}$, and a proxemic radius $r_{\mathrm{prox}}$ according to Hall's proxemic theory \cite{HallThehiddendimension1990}.
The world state $\bm{s}_t = [\bm{s}_{t}^0, \cdots, \bm{s}_{t}^{N-1}]$ represents the environment at time $t$.
One episode's trajectory $\tau$ is the sequence of states, observations, actions, and rewards within the terminal time $T$.
The return of one episode $\mathcal{R}(\tau) = \sum_{t=0}^T \gamma^t R_t$ is the accumulated and discounted reward $R_t$.
The central objective is to learn the optimal robot policy $\pi^*$, which maximizes the expected return:
\begin{align}
    \mathcal{T}(\tau|\pi) &= \mathcal{T}_0 \prod_{t=0}^T \mathcal{T}(\bm{s}_{t+1}|\bm{s}_t, \bm{a}_t) \pi(\bm{a}_t|\bm{o}_t)\Omega(\bm{o}_t|\bm{s}_t)\text{,}\\
    \underset{\tau \sim \pi}{\mathbb{E}} [\mathcal{R}(\tau)] &= \int_\tau \mathcal{T}(\tau|\pi)\mathcal{R}(\tau) \text{,}\\
    \pi^*(\bm{a}|\bm{o}) &=  \arg \underset{\pi}{\max} \underset{\tau \sim \pi}{\mathbb{E}} [\mathcal{R}(\tau)]\text{.}
\end{align}
Considering a stochastic environment, $\mathcal{T}(\tau|\pi)$ is the probability of a trajectory starting in $\bm{s}_0$ with $\mathcal{T}_0$.

We follow the framework in \cite{FloegelSociallyIntegratedNavigation2024} to train a \acf{SI} navigation policy and use the same reward and observation system.
The policy is trained from scratch, respects proxemic and velocity social norms, and exhibits social behavior that is adaptive to individual human preferences.
The action of the policy is a velocity $v_t$ and a delta heading $\Delta \theta_t$ command.

\section{Approach}
This section first introduces \acf{ODV} into the \ac{PPO} algorithm to enable aleatoric uncertainty estimates. 
Subsequently, we describe aleatoric, epistemic, and predictive uncertainty estimation in the \ac{DRL} policy using \ac{MC-Dropout} and deep ensembles. 
Finally, we provide a \acf{POC} estimation based on the uncertainty estimates and change the robot's navigation strategy into cautious \ac{SCA} for risk-aware action selection in novel and perturbed environments.

\subsection{Observation-Dependent Variance}
In \ac{PPO}\cite{SchulmanProximalPolicyOptimizationAlgorithms2017}, the actor outputs multivariate Gaussian distributed actions $\mathcal{N}(\bm{\mu}_{\mathrm{a}}(\bm{o}_t),\bm{\sigma}^2_{\mathrm{a}})$ with an observation dependent mean but a parameterized variance, independent from observation. 
An observation-independent variance does not allow for aleatoric uncertainty estimation of the action selection process\cite{FanLearningResilientBehaviorsforNavigat2020}.
Therefore, the actor network is adapted, and a linear layer is incorporated to output the observation-dependent variances for the actions as $\log(\bm{\sigma}_{\mathrm{a}})$.
This leads to a multivariate Gaussian distributed policy $\mathcal{N}(\bm{\mu}_{\mathrm{a}}(\bm{o}_t),\bm{\sigma}^2_{\mathrm{a}}(\bm{o}_t))$ where both the mean and variance of the policy depend on the observation.
However, using \ac{ODV} leads to stability problems during training.
First, the variance can be arbitrarily large, which can cause the normal distribution to lose its shape, become a uniform distribution, and disrupt learning because actions are sampled randomly.
Second, policy updates can lead to high variances at the late stages of training, which, by sampling, may lead to a sequence of poor actions, a series of bad updates, and, finally, performance collapse.
We propose to use variance clamping for the first problem to prevent the deformation of the normal distribution by limiting the maximum variance.
A grid search for the clamping values was done, and the highest maximum clamping value that did not cause collapse and still resembled a normal distribution was used.
The second problem is addressed by modifying the \ac{PPO} loss by adding a \ac{MSE} loss for the action variance
\begin{equation}\label{eq:MSE_variance_loss}
    \mathcal{L}_{\mathrm{\theta}}^{\mathrm{\sigma}} \left(\pi\right) =  \frac{t}{T} \cdot \lambda_{\mathrm{\sigma}} \cdot\frac{1}{B} \sum_{i=1}^{B}\frac{1}{2}\cdot 
    \left\Vert \bm{\sigma}^2_{\mathrm{a}} - \bm{\sigma}^2_{\mathrm{tgt}}\right\Vert^2 \text{,}
\end{equation}
where $\frac{t}{T}$ accounts for exploration-exploitation trade-off with the current timestep $t$ and the total training timesteps $T$ and batch size of $B$.
The constant $\lambda_{\mathrm{\sigma}}$ scales and weights the variance loss, and $\bm{\sigma}^2_{\mathrm{tgt}}$ is a desired target variance for each action.
This leads to total policy loss $\mathcal{L}_{\mathrm{\theta}}(\pi) = \mathcal{L}_{\mathrm{\theta}}^{\mathrm{CLIP}}(\pi) + \mathcal{L}_{\mathrm{\theta}}^{\mathrm{\sigma}}(\pi)$
where $\mathcal{L}_{\mathrm{\theta}}^{\mathrm{CLIP}}(\pi)$ is the PPO clipping loss.
We refer to the full model as \ac{ODV-PPO}.

\subsection{Uncertainty Estimation}
\begin{figure}[!tb]
    \vspace{0.5em}
    \centering
    \includegraphics[width=0.90\linewidth]{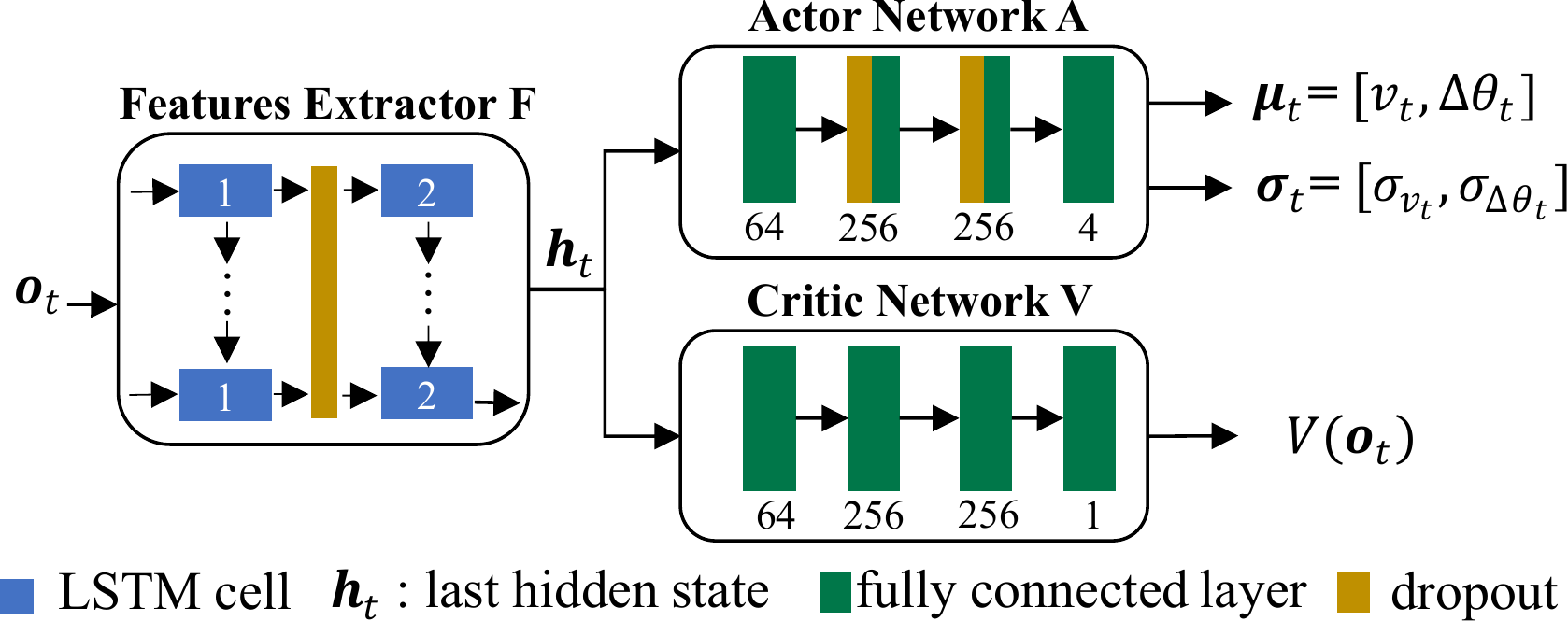}
    \caption{Our \ac{ODV-PPO} network, a common actor-critic architecture, with a shared features extractor and separated actor and value networks. The actor and the features extractor contain dropout layers. The actor outputs an observation-dependent mean and variance.}
    \label{fig:network_dropout}
    \vspace{-1.5em}
\end{figure}

\subsubsection{MC-Dropout}

We incorporate dropout to approximate Bayesian inference in deep Gaussian processes, referred to as \ac{MC-Dropout} \cite{GalDropoutasaBayesianApproximation:Rep2015}.
Implemented in two parts of the \ac{ODV-PPO} network, in the $2$-layer LSTM features extractor and in the hidden layers of the actor-network, as depicted in Fig.~\ref{fig:network_dropout}. 
The critic network does not contain a dropout to avoid instabilities in the target value \cite{LockwoodAReviewofUncertaintyforDeepReinfor2022}.
The policy is first trained with dropout probability $p_{\mathrm{train}}$.
In the testing phase, the samples are drawn with a higher dropout $p_{\mathrm{test}}$ to estimate the uncertainty.
We sample $K$ independent policy predictions with dropout of the action distribution $\bm{\mu}_k,\bm{\sigma}_k = \pi(\cdot|\bm{o}_t)$ for an observation at time step $t$. 
Subsequently, we calculate the uncertainties as proposed in \cite{charpentier_disentangling} from the individual samples.
The epistemic uncertainty is estimated with the variance of the means with
\begin{equation}\label{eq:epistemic}
    \bm{u}^{\mathrm{ep}} = \mathbb{V}[\bm{\mu}] \approx \frac{1}{K}\sum_{k=1}^K (\bm{\mu}_k - \overline{\bm{\mu}}_k)^2 \text{,}
\end{equation}
where $\overline{\bm{\mu}}_k$ is the average of the mean predictions obtained with dropout at time $t$.
Additionally, we calculate the mean over the variances, which represents the aleatoric uncertainty
\begin{equation}\label{eq:aleatoric}
    \bm{u}^{\mathrm{al}} = \mathbb{E}[\bm{\sigma}^2]   \approx \frac{1}{K}\sum_{k=1}^K \bm{\sigma}^2_k \text{.}
\end{equation}
Both uncertainty estimates are vectors with the dimension of the action space.

\subsubsection{Deep Ensembles}

For deep ensembles, $K$ networks are trained separately with different weight initialization and in environments with different seeds. 
A single network in the ensemble has the same architecture as in Fig.~\ref{fig:network_dropout} and is trained in the same way as for \ac{MC-Dropout}. 
The dropout is used in the training for regularization but not used to calculate uncertainty estimates.
In testing, each model of the ensemble gets the same observation.
Thus, we generate $K$ independent samples of the mean and variance of the actions. 
The epistemic uncertainty is then calculated with \eqref{eq:epistemic} and the aleatoric with \eqref{eq:aleatoric}.

\subsubsection{Features Extractor Uncertainty}
In addition to the epistemic and aleatoric uncertainty in the actor network, we estimate the predictive uncertainty of the features extractor, which does not distinguish between aleatoric and epistemic uncertainty.
We distinguish between the predictive features uncertainty estimation with \ac{MC-Dropout} and with the deep ensembles method.

For \ac{MC-Dropout}, we get $K$ samples of the hidden state $\bm{h}_t$ at each time step $t$ to estimate the uncertainty. 
Based on the $K$ features vectors, we estimate a degree of uncertainty based on the method proposed in \cite{Kim_mapping} and based on the element-wise variance in the features vector $\bm{h}_t$ at each time step. 
This features variance vector is
\begin{equation}\label{eq:LSTM_dim_problem}
    \mathbb{V}(\bm{h}) = \begin{bmatrix}
           \mathbb{V}[h_1] \\
           \vdots \\
           \mathbb{V}[h_{D}]
         \end{bmatrix} =  \begin{bmatrix}
           \frac{1}{K}\sum_{i=1}^K\left(h_{1,i} - \overline{h}_1\right)^2 \\
           \vdots \\
           \frac{1}{K}\sum_{i=1}^K\left(h_{D,i} - \overline{h}_{D}\right)^2
         \end{bmatrix} 
         \text{,}
\end{equation}
where $D$ is the dimension of $\bm{h}$ and $h_{d,i}$ is the $d$-th features vector element of the $i$-th sample from the $K$ forward passes, and $\overline{h}_d = \frac{1}{K}\sum_{i=1}^K h_{d,i}$ is the mean for each element over the samples.
During training, the minimum and maximum of each element in the features vector are tracked as $h_d^{\mathrm{min}}$ and $h_d^{\mathrm{max}}$.
Based on these values, an upper bound of the variance $\mathbb{V}[h_d]^{\mathrm{max}} = \frac{\left(h^{\mathrm{max}}_d-h^{\mathrm{min}}_d\right)^2}{12}$ is calculated for each element, which occurs when the distribution follows a uniform distribution.
Since the dimension of the features vector is commonly very high, a mapping function $g(\cdot): \mathbb{R}^d \rightarrow \mathbb{R} $ is used to map the variance to a degree of uncertainty $u^{\mathrm{feat}}$ as proposed in \cite{Kim_mapping}:
\begin{equation} \label{eq:feat_unc_dropout}
    u^{\mathrm{feat}} = g(\mathbb{V}(\bm{h})) = \sum_{d=1}^D \frac{\mathbb{V}({h}_d)}{\sum_{d=1}^{D} \mathbb{V}({h}_d)} \frac{\mathbb{V}({h}_d)}{\mathbb{V}[h_d]^{\mathrm{max}}} \text{,}
\end{equation}    
where $\mathbb{V}({h}_d)$ is the $d$-th element of the features variance vector and $\mathbb{V}[h_d]^{\mathrm{max}}$ is used to normalize the uncertainty.

For deep ensembles, the uncertainty mapping is performed with the average of the features uncertainty
\begin{equation} \label{eq:feat_unc_ensemble}
    u^{\mathrm{feat}} = g(\mathbb{V}(\bm{h})) = \frac{1}{D}\sum_{d=1}^D\mathbb{V}({h}_d) \text{.}
\end{equation}
Using \eqref{eq:feat_unc_dropout} would require training the ensemble models in parallel in the same environment, which is computationally heavy for a large $K$.

\begin{table}[!b]
    \centering
    \vspace{-0.5em}
    \caption{ODV-PPO Hyperparameters}
    \vspace{-0.5em}
    \label{tab:experiment_parameters}
    \begin{tabular}{l c || l c}
        \hline
        Parameter & Value & Parameter & Value \\
        \hline
        variance factor $\lambda_{\mathrm{\sigma}}$ & \num{0.3}     &   dropout rate $p_{\mathrm{train}}$ & \num{0.1}       \\
        target variance $\bm{\sigma}^2_{\mathrm{tgt}}$ & $\bm{0}$   & dropout rate $p_{\mathrm{test}}$ & \num{0.5} \\
        variance clamping min & \num{-20}                           &   samples/models $K$ & \num{20}    \\
        variance clamping max & \num{0.25}                          &  learning rate & \num{0.00025}   \\
        batch size & \num{128}                                      & num steps & \num{128}   \\
        
        \hline
    \end{tabular}
\end{table}

\begin{figure*}[!ht] 
    \centering
    \vspace{0.4em}
    \includegraphics[width=0.99\linewidth]{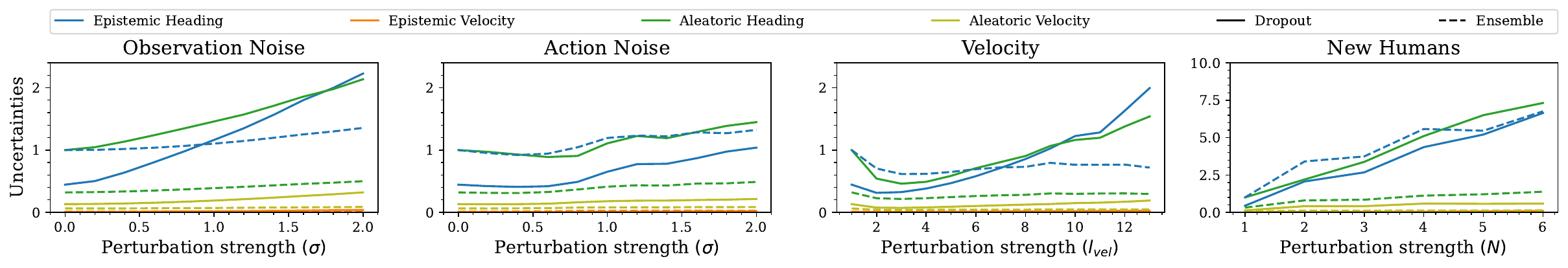}
    \vspace{-0.5em}
    \caption{Normalized and episodic mean values of epistemic and aleatoric uncertainty estimation using the \ac{MC-Dropout} and deep ensemble approach.}
    \label{fig:epistemic_aleatoric_uncertainty}
    \vspace{-1.5em}
\end{figure*}

\subsection{Uncertainty-Aware Action Selection}
To avoid collision in scenarios where the robot is uncertain in action selection, we use the uncertainty estimates for a \ac{POC} estimation. 
For \ac{POC} estimation, we investigated the usability of a sigmoid function, a \ac{PCN}, and a threshold function, all with uncertainties as input. 
The sigmoid function could not be parameterized manually.
The \ac{PCN}, inspired by \cite{B.LutjensSafeReinforcementLearningWithModelU2019} additionally considers the temporal evolution and is trained in a supervised manner based on gathered \ac{DRL} rollouts. 
However, the \ac{PCN} only performed well for action noise in our experiments, and we could not claim the generalization of the \ac{PCN} itself therefore, we decided on a white box approach for the \ac{POC} since the risk assessment is an essential part for safe action selection. 
Thus, we use the threshold function 
\begin{equation}\label{eq:prob_col_combined}
    \mathbb{P}(\mathrm{col}) = 
    \begin{cases}
        1, & \text{if } (c_{\mathrm{ep}} \lor c_{\mathrm{feat}}) \land c_{\mathrm{prox}} \land c_{\mathrm{ap}}  \\
        0, & \text{otherwise}\quad
    \end{cases}
    \text{,}
\end{equation}
which indicates if the uncertainty condition, distance analysis, and scenario measure turn true. 
The uncertainty conditions indicate if the heading epistemic $ c_{\mathrm{ep}} = \mathds{1}[{u}^{\mathrm{ep}}_{\mathrm{head}} > \lambda_{\mathrm{ep}}] $ or the predictive feature uncertainty $c_{\mathrm{feat}} = \mathds{1}[u^{\mathrm{feat}} > \lambda_{\mathrm{f}}]$ are beyond a threshold. 
To analyze the trend of uncertainty in the trajectory, we smooth the noisy step-wise uncertainty estimates by windowing them over $w$ previous steps.
The proximity condition flags $c_{\mathrm{prox}} = \mathds{1}[(d^{0,j}_t + \beta_1 \cdot |\Delta v^{0,j}_t|)< \lambda_{\mathrm{prox}}]$ scenarios with the close distance $d^{0,j}_t$ to another human and takes into account the relative velocity $\Delta v^{0,j}_t$ to the nearest human. 
The approach condition $c_{\mathrm{ap}} = \mathds{1}[d^{0,j}_{t-1}\leq d^{0,j}_{t}]$ considers that once the robot is moving away from the human, the risky scenario was averted.

With the \ac{POC}, a risk-aware action selection can be proposed by either reducing the velocity or using a cautious policy when the \ac{POC} is high.
Since reducing the velocity is not always safe when humans move fast \cite{B.LutjensSafeReinforcementLearningWithModelU2019}, we use a cautious policy. 
According to the social navigation taxonomy in \cite{FloegelSociallyIntegratedNavigation2024}, we propose to change the robot's interaction behavior to a \ac{SCA} strategy in high-risk scenarios instead of a \ac{SA} or \ac{SI} strategy and focus on physical collision avoidance in the first regard. 
As a collision avoidance strategy, we use the \ac{ORCA} \cite{vandenBergReciprocalnBodyCollisionAvoidance2011} model.
To increase the caution, the radius of the surrounding humans is assumed to be $1.5$ times bigger as observed, promoting trajectories away from humans and considering a safety zone.

\section{Evaluation}
\begin{table*}[!ht]
\tiny
\centering
\vspace{0.4em}
\caption{Evaluation results across 15 seeds with 200 episodes per scenario and seed. Goal, Return: $\uparrow$; Coll., TO (timeout), PV (proxemic violation) $\downarrow$}
\vspace{-0.5em}
\setlength{\tabcolsep}{3.5pt}
\begin{tabular}{|c|c||c|c|c|c|c||c|c|c|c|c||c|c|c|c|c|c|c|}
\hline
\multirow{2}{*}{\makecell{Training \\ Scenario }} & \multirow{2}{*}{Approach} &   \multicolumn{5}{c||}{Eval. Scenario: Circle Interaction}        & \multicolumn{5}{c||}{Eval. Scenario: Circle Crossing}        & \multicolumn{5}{c|}{Eval. Scenario: Random} \\ \cline{3-17}
                                                &                       & Goal (\%)    & Col. (\%)     & TO (\%)    & PV & Return ($\mu$ $\pm$ $\sigma$) & Goal (\%)    & Col.(\%)     & TO (\%)       & PV  & Return ($\mu$ $\pm$ $\sigma$) & Goal (\%)    & Col. (\%)    & TO (\%)         & PV  & Return ($\mu$ $\pm$ $\sigma$) \\ \hline \hline
\multirow{4}{*}{\makecell{Circle \\ Interaction }}  & SI \cite{FloegelSociallyIntegratedNavigation2024} (PPO)  & $99.80$      & $0.20$        &$ 0.00$     &  $7$       & $4.5 \pm 0.46$                 & $93.30$      & $ 4.37$      & $2.33$       & $169$      & $3.66 \pm 2.09$               & $76.80$      &  $4.70$      & $18.50$         & $43$   & $2.63 \pm 3.63$ \\
                                                    & PPO-Drop          & $99.90$      & $0.10$        &$ 0.00$     &  $8$       & $\bm{4.66} \pm \bm{0.42}$      & $94.40$      & $ 3.83$      & $1.77$       & $200$      & $3.92 \pm 2.01$               & $78.70$      &  $5.40$      & $15.90$         & $57$   & $2.93 \pm 3.65$ \\
                                                    & ODV-PPO no Loss   & $99.47$      & $ 0.53$       & $ 0.00$    &  $18$      & $4.25 \pm 0.65$                & $88.57$      & $ 8.73$      & $ 2.70$      &  $195$     & $2.94 \pm 2.64$               & $79.27$      &  $7.37$      & $13.37$         &  $69$  & $2.22 \pm 3.59$ \\
                                                    & ODV-PPO           & $99.23$      & $ 0.77$       & $ 0.00$    &  $17$      & $4.29 \pm 0.78$                & $89.93$      & $ 7.80$      & $ 2.57$      &  $225$     & $3.12 \pm 2.55$               & $77.77$      &  $7.50$      & $14.73$         & $80$   & $2.15 \pm 3.79$ \\
                                                    & ODV-PPO-Drop      & $99.90$      & $ 0.10$       & $ 0.00$    &   $\bm{1}$  & $4.52 \pm 0.36$                & $\bm{96.87}$ &  $\bm{2.47}$ &  $\bm{0.67}$ &  $\bm{101}$  & $\bm{4.04} \pm \bm{1.45}$     & $\bm{82.00}$ &  $\bm{4.50}$ & $\bm{13.50} $ & $\bm{40}$   & $\bm{3.02} \pm \bm{3.39}$ \\ \hline \hline

\multirow{4}{*}{\makecell{Circle \\ Crossing }}     & SI \cite{FloegelSociallyIntegratedNavigation2024} (PPO)             & $49.57$      & $12.20$      & $38.23$     &  $101$     &  $-1.25 \pm 5.16$               & $86.10$      & $7.07$      & $6.83$       &  $52$      &  $2.17 \pm 3.49$             &$ 58.90$       & $7.83$       & $33.27$         & $90$   &  $-0.57 \pm 5.39$ \\
                                                    & PPO-Drop          & $87.80$      & $ 1.13$      & $11.07$     &  $53$      &  $3.15 \pm 2.78$                & $98.13$      & $1.57$      & $0.30$       &  $64$      &  $3.91 \pm 1.24$             &$ 82.33$       & $4.07$       & $13.60$         & $53$   &  $2.71 \pm 3.61$ \\                                                 
                                                    & ODV-PPO no Loss   & $41.57$      & $ 9.43$      & $49.00$     &  $147$    &  $-2.28 \pm 4.98$               & $71.47$      & $ 2.03$     & $26.50$       &  $63$      &  $0.83 \pm 4.45$              & $40.80$      & $8.63$       & $50.57$         &  $95$  &  $-2.74 \pm 5.3$ \\
                                                    & ODV-PPO           & $38.37$      & $ 6.67$      & $54.97$     &  $76$     &  $-2.24 \pm 5.42$               & $58.90$      & $ 8.70$     & $32.40$       &  $87$     &  $0.18 \pm 4.84$               & $39.60$      & $7.17$       & $53.23$         &  $66$  &  $-2.51 \pm 5.87$ \\
                                                    & ODV-PPO-Drop      & $\bm{92.90}$ &  $\bm{0.67}$ & $\bm{6.43}$ & $\bm{21}$  &$\bm{3.78} \pm \bm{2.35}$       & $\bm{99.10}$ &  $\bm{0.83}$ &  $\bm{0.07}$ &  $\bm{35}$  &  $\bm{4.22} \pm \bm{0.88}$      & $\bm{88.07}$ & $\bm{2.10}$  &  $\bm{9.83}$  & $\bm{49}$  &  $\bm{3.46} \pm \bm{3.14}$ \\ \hline \hline           

\multirow{4}{*}{Random}                             & SI \cite{FloegelSociallyIntegratedNavigation2024} (PPO)             & $68.03$      &  $5.60$      & $26.37$     &  $159$   & $1.36 \pm 4.39$                  & $68.40$      & $ 8.60$      & $23.00$    & $179$  & $1.15 \pm 4.12$                      & $80.40$      & $ 2.37$      & $17.23$       & $66$ &  $2.44 \pm 3.80$ \\
                                                    & PPO-Drop          & $91.67$      &  $2.43$      & $ 5.90$     &  $67$    & $3.4 \pm 2.59$                   & $86.70$      & $ 8.13$      & $ 5.17$     & $219$  & $2.76 \pm 3.12$                    & $90.20$       & $ 3.30$      & $ 6.50$       & $64$   &  $3.28 \pm 2.86$ \\
                                                    & ODV-PPO no Loss   & $ 0.00$      &  $0.37$      & $99.63$     &  $11$    & $-6.53 \pm 0.37$                 & $ 0.00$      & $ 0.10$      & $99.90$      &   $11$  & $-6.54 \pm 0.71$                 & $ 0.00$       & $ 0.10$      & $99.90$       &  $6$  &  $-6.39 \pm 0.80$ \\
                                                    & ODV-PPO           & $58.50$      &  $2.20$      & $39.30$     &  $73$    & $0.19 \pm 4.81$                  & $53.53$      & $ 3.30$      & $43.17$      &   $112$  & $-0.26 \pm 4.62$                & $59.53$       & $ 1.17$      & $39.30$       & $47$   &  $0.36 \pm 4.80$ \\
                                                    & ODV-PPO-Drop      & $\bm{99.53}$ &  $\bm{0.30}$ & $\bm{ 0.17}$& $\bm{8}$ & $\bm{4.33} \pm \bm{0.74}$        & $\bm{97.53}$ &  $\bm{2.43}$ &  $\bm{0.03}$ &   $\bm{77}$  & $\bm{4.04} \pm \bm{1.33}$        & $\bm{99.43}$  & $\bm{0.47}$ & $\bm{0.10}$     & $\bm{14}$   &  $\bm{4.39} \pm \bm{0.83}$ \\                                             
 \hline

\end{tabular}
\label{tab:training_evaluation}
\vspace{-2.0em}
\end{table*}
We follow the principles and guidelines of evaluating social navigation \cite{FrancisPrinciplesandGuidelinesforEvaluating2023} and evaluate the algorithm's response to dynamic obstacles in a reproducible simulation environment as a prior step of real-world experiments. 
First, we compare the training and generalization performance of the proposed \ac{ODV-PPO} algorithm in different scenarios. 
Subsequently, we compare the capability of \ac{MC-Dropout} and deep ensembles to disentangle different types of perturbations.
Finally, we investigate the suitability of the uncertainty estimates to detect risky scenarios and avoid collisions. 

\subsection{Experimental Setup}
We use the microscopic and social-psychological simulation environment from \cite{FloegelSociallyIntegratedNavigation2024} since the \ac{ORCA} motion model is representative of human behavior \cite{Samavi_SICNav} and the augmentation with human social interaction behavior allows us to train an \ac{SI} policy.
\textit{Stables Baselines3} \cite{AntoninRaffinStableBaselines3ReliableReinforcemen2021} is used to train and evaluate all policies. 
For \ac{PPO} with and without dropout, we use the optimized hyperparameters from \cite{FloegelSociallyIntegratedNavigation2024}. 
For all \ac{ODV-PPO} variants, we adapted the hyperparameters based on an Optuna \cite{optuna_2019} hyperparameter search, as stated in Table~\ref{tab:experiment_parameters}.

\begin{figure}[!b]
    \centering
    \vspace{-1.5em}
    \begin{subfigure}[b]{.21\textwidth}
      \centering
      \includegraphics[width=0.9\linewidth]{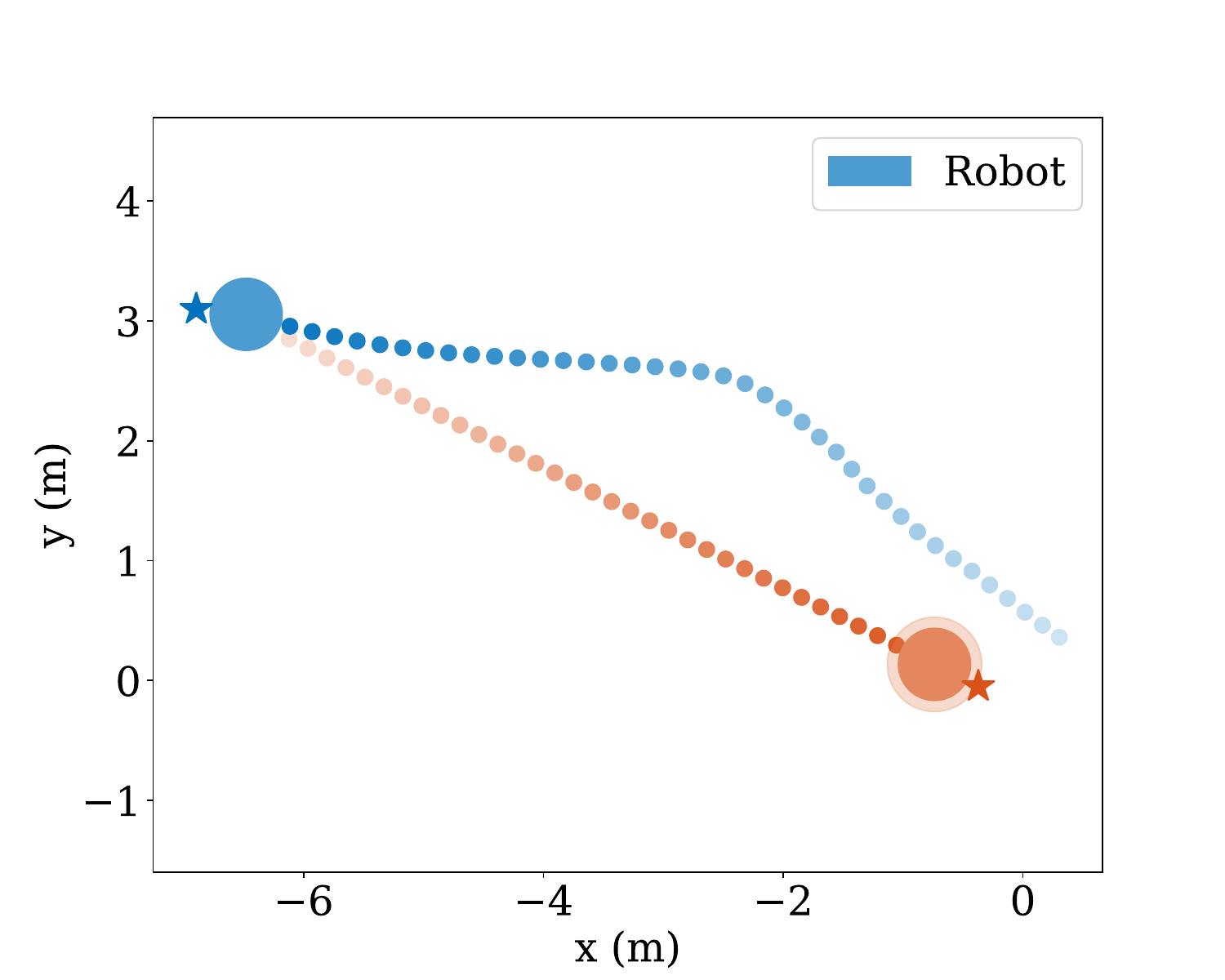}
    \end{subfigure}
    \begin{subfigure}[b]{.26\textwidth}
      \centering
      \includegraphics[width=0.9\linewidth]{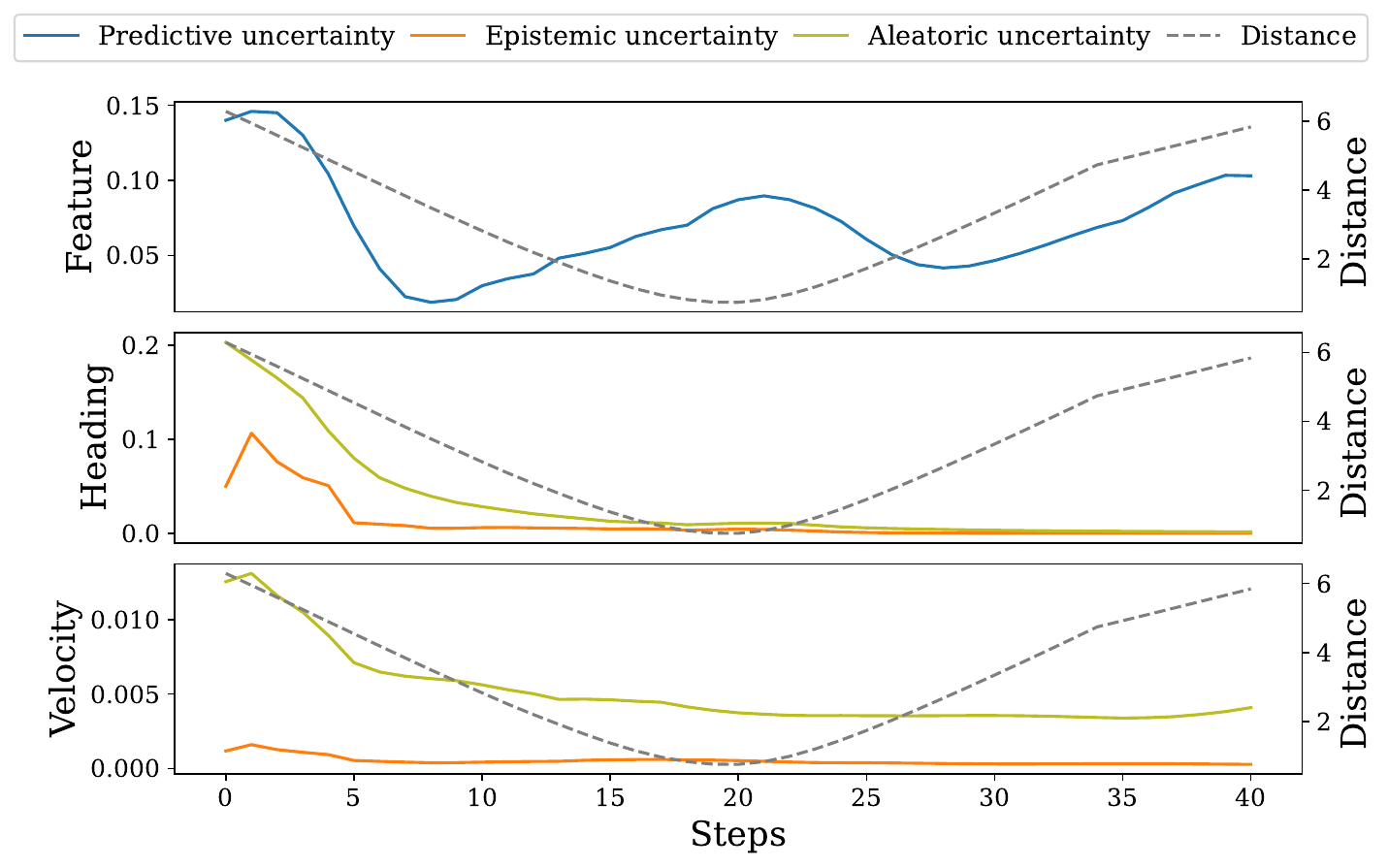}
    \end{subfigure}
    
    \caption{Windowed uncertainty estimates ($w=4$) using \ac{MC-Dropout} approach. Position swap scenario with start and goal position perturbation.}
    \label{fig:stepwise_uncertainty}
    \vspace{-0.2em}
\end{figure}

To evaluate the design choices of the \ac{ODV-PPO} and the impact of dropout, we consider three different scenarios: a circle-interaction \cite{FloegelSociallyIntegratedNavigation2024}, a circle crossing, and a random scenario. 
In all scenarios, the agents' preferred speed and humans' proxemic radius are sampled from $\mathcal{U}[0.5, 1.0]$ and $\mathcal{U}[0.3, 0.7]$, respectively. 
The start and goal positions are randomly sampled within the red areas, as depicted in Fig.~\ref{fig:training_results_ODV_PPO}, and the robot is randomly assigned to one of the blue start-goal pairs. 
We train $15$ policies per algorithm across different seeds in each scenario and evaluate each policy for $200$ episodes in all scenarios to investigate the generalization. 

We train new policies for uncertainty estimation and consider a position swap scenario where the human position is randomly sampled on a circle with $7$\,m radius.
All policies are trained in this simple position swap scenario with one human, and the human proxemic radius is sampled with $r_{\mathrm{prox}}=\mathcal{U}(0.3,0.4)$.
To analyze the uncertainty estimates, the position swap scenario is systematically expanded with perturbations, noise, and more humans.
We stimulate aleatoric uncertainty with observation and action noise, and epistemic uncertainty with an increased velocity of humans and an increased number of humans.
Observation noise is modeled with additive Gaussian noise $\mathcal{N}\left(\bm{0},I\cdot\sigma_{\mathrm{obs}}^2\right)$.
Action noise for the heading is modeled with additive Gaussian noise $\mathcal{N}\left(0,\sigma_{\mathrm{head}}^2\right)$ and the velocity is scaled with a uniformly sampled factor $ \mathcal{U}\left(1-\sigma_{\mathrm{vel}},1\right)$ with $\sigma_{\mathrm{vel}}\leq1$ to simulate terrain grip and avoid actions with negative velocity.
The epistemic uncertainty is introduced through scaling the preferred velocity $l_{\mathrm{vel}} \cdot v_{\mathrm{pref}}^i$ of humans and adding multiple humans into the environment to extend the position swap scenario into a circle-crossing scenario.

\subsection{Results}

\subsubsection{Ablation and Generalization}

\begin{figure}[!ht]
    \centering
    \begin{subfigure}[b]{.33\textwidth}
      \centering
      \includegraphics[width=\linewidth]{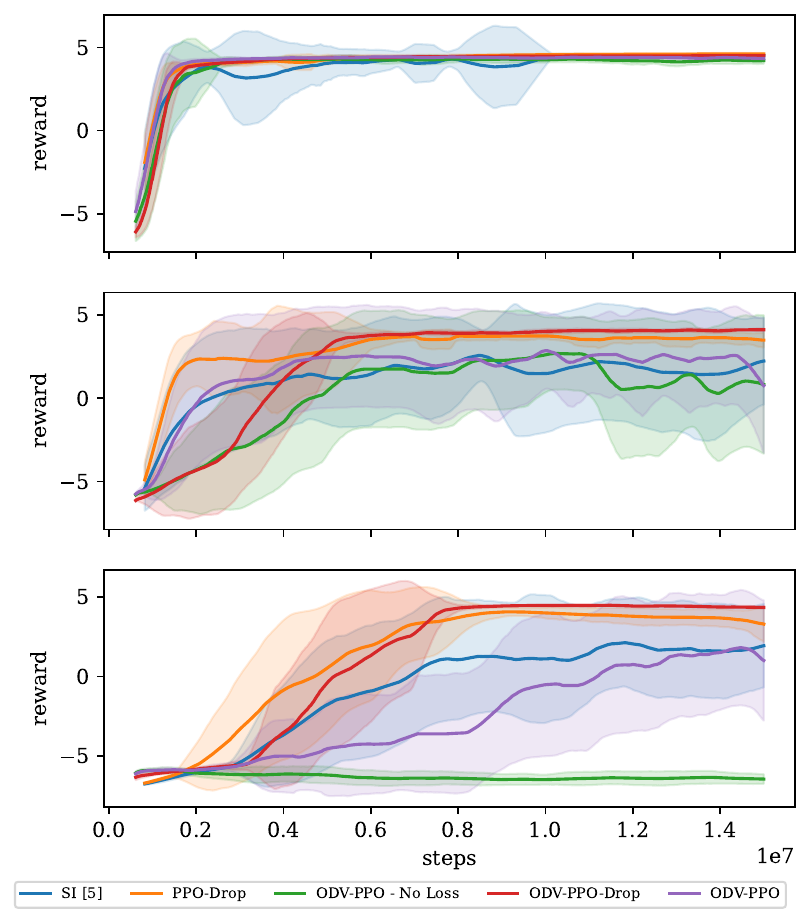}
    \end{subfigure}
    \begin{subfigure}[b]{.11\textwidth}
      \centering
      \raisebox{2.4em}{\includegraphics[width=0.88\linewidth]{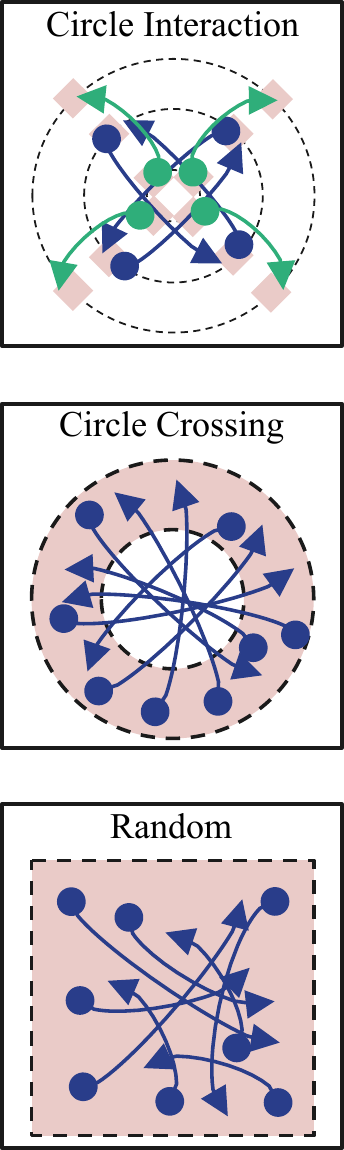}}
    \end{subfigure}
    \vspace{-0.5em}
    \caption{Training reward across $15$ seeds for different algorithms and scenarios. The agent's goal and start positions are sampled in red areas.}
    \label{fig:training_results_ODV_PPO}
\end{figure}

The training rewards in Fig.~\ref{fig:training_results_ODV_PPO} and aggregated evaluation results in Table~\ref{tab:training_evaluation} show that naively adding \ac{ODV} into \ac{PPO} does not lead to good results or even to non-convergence. 
The newly introduced loss function in \eqref{eq:MSE_variance_loss} stabilizes the training convergence behavior, but the evaluation results are worse than for \ac{PPO} only, as used in \cite{FloegelSociallyIntegratedNavigation2024}. 
In all scenarios, \ac{PPO} with dropout improves training rewards and the generalization performance to out-of-distribution scenarios (the policy is not trained on).
However, the combination of dropout and \ac{ODV-PPO} leads to the best training rewards and evaluation results except for the circle interaction in-distribution scenario evaluation (same training and evaluation scenario) for the circle-interaction scenario. 
In all scenarios, \ac{ODV-PPO} with dropout shows the best generalization performance to out-of-distribution scenarios, has the highest success rate (reached goal), and has the least collisions.
All policies trained in the circle-interaction scenario converge very fast, have low variance throughout the seeds, and have good generalization results compared to those trained on circle crossing or random scenarios. 
\ac{ODV} allows the policy to adjust the exploration-exploitation trade-off for each observation individually. 
Thus, the policy can explore more in human-robot interaction scenarios, which is challenging for a parametrized variance. 
Despite higher rewards, this is also reflected in the lower proxemic violations throughout all runs as stated in Table~\ref{tab:training_evaluation}.

\begin{figure*}[!htb] 
    \centering
    \vspace{0.4em}
    \includegraphics[width=\linewidth]{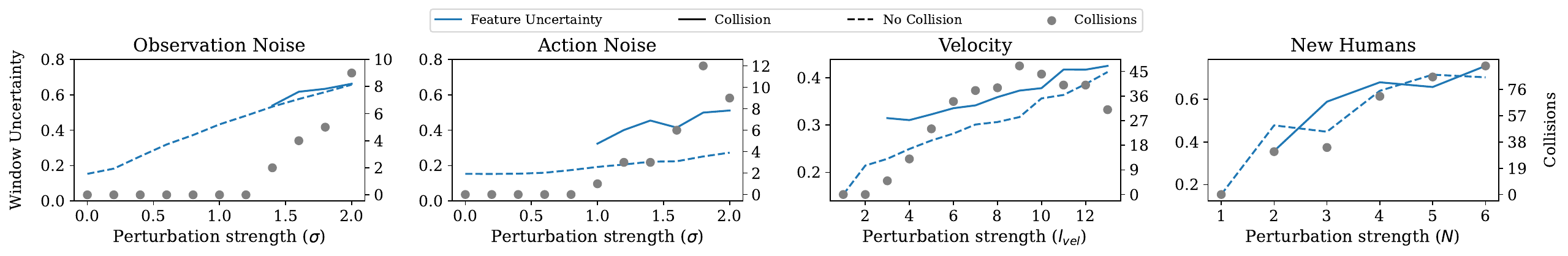}
    \vspace{-1.5em}
    \caption{Windowed predictive uncertainty of features extractor using \ac{MC-Dropout} approach on step before collision or at closest distance.}
    \label{fig:perturbations_and_uncertainties}
    \vspace{-1.5em}
\end{figure*}

\subsubsection{Disentangling Uncertainty Estimation}
We evaluate the \ac{MC-Dropout} and Deep Ensemble approach with different perturbation sources and strengths and estimate the uncertainties in each step. 
Observation and action noise is $\sigma_{\mathrm{head}}, \sigma_\mathrm{vel}, \sigma_{\mathrm{obs}}\in [0,0.2,\cdots,2]$, velocity scaling factor is $l_{\mathrm{vel}} \in[1,..,8]$, and new humans are $N\in[2,..,7]$.
The estimated aleatoric and epistemic uncertainties are depicted in Fig.~\ref{fig:epistemic_aleatoric_uncertainty}. 
For both approaches, the results show that the velocity uncertainty is low and flat compared to the heading uncertainty, which increases approximately linearly, and new humans cause the highest uncertainty. 
In addition, the \ac{MC-Dropout} approach is capable of disentangling the source of the uncertainty, but we also observed that a proper dropout rate $p_{\mathrm{test}}$ is crucial for this capability.
In contrast, the deep ensemble approach always has high epistemic uncertainty and cannot distinguish between aleatoric and epistemic uncertainty. 
The results are also aggregated in Table~\ref{tab:perturbations_uncertainties_and_methods} for the normalized rate of change of the uncertainties for the perturbation.
The normalized value is $\frac{\Delta U}{\Delta\sigma}=\frac{1}{u_\mathrm{\sigma}(0)}\frac{u_{\mathrm{\sigma}(\mathrm{max})}-u_{\mathrm{\sigma}(0)}}{\sigma(\mathrm{max})-\sigma(0)}$ with the maximum perturbation strength $\sigma(\mathrm{max})$, the mean uncertainty $u_{\mathrm{\sigma}(\mathrm{\mathrm{max}})}$ per strength, and the mean uncertainty $u_{\mathrm{\sigma}(0)}$ in the unperturbed environment of the episodes.

To analyze the trend of uncertainty as the robot approaches and interacts with the human, the mean uncertainty $\overline{u}_w$ over a window of the last $w=4$ steps is evaluated. 
The experiments show that the predictive uncertainty increases for both approaches in close situations, as exemplified in Fig.~\ref{fig:stepwise_uncertainty}, but the action uncertainty is low.
For \ac{MC-Dropout} approach, the windowed predictive feature uncertainty is one step before collision (if occurred) higher for action and velocity noise compared to the step of minimum distance (if no collision occurred), as depicted in Fig.~\ref{fig:perturbations_and_uncertainties}. 
However, there is no distinction for the deep ensemble approach, which is additionally less sensitive. 
In addition, the results show that the number of collisions increases with the perturbation strength, as depicted with grey dots in Fig.~\ref{fig:perturbations_and_uncertainties}.

\begin{table}[!htb]
\setlength{\tabcolsep}{3pt}
\tiny
\centering
\vspace{0.5em}
\caption{Normalized aggregated uncertainty estimates.}
\vspace{-0.5em}
\setlength{\tabcolsep}{3pt}
    \begin{tabular}{|c|c|c|c|c|c|c|c|c|}
    \hline
    \multirow{2}{*}{{Uncertainties}} & \multicolumn{2}{c|}{Observation Noise} & \multicolumn{2}{c|}{{Action Noise}} & \multicolumn{2}{c|}{{Velocity}} & \multicolumn{2}{c|}{{New Humans}} \\ \cline{2-9}
                           & Dropout & Ensemble & Dropout & Ensemble & Dropout & Ensemble & Dropout & Ensemble \\ \hline \hline
    \makecell{Predictive \\ Feature}  & \textbf{1.84} & 0.11 & \textbf{0.87} & 0.03 & \textbf{0.20} & 0 & \textbf{0.65} & 0.38 \\ \hline
    \makecell{Aleatoric \\ Heading}   & \textbf{0.57} & 0.28 & 0.22 & \textbf{0.26} & \textbf{0.04} & 0 & \textbf{0.90} & 0.47 \\ \hline
    \makecell{Epistemic \\ Heading}  & \textbf{2.00} & 0.18 & \textbf{0.67} & 0.16 & \textbf{0.25} & 0 & \textbf{1.99} & 0.82 \\ \hline
    \makecell{Aleatoric \\ Velocity}  & \textbf{0.71} & 0.19 & \textbf{0.31} & 0.18 & \textbf{0.03} & 0 & \textbf{0.49} & 0.17 \\ \hline
    \makecell{Epistemic \\ Velocity} & \textbf{4.08} & 0.49 & \textbf{1.46} & 0.64 & \textbf{0.39} & 0.02 & \textbf{3.01} & 0.30 \\ \hline
    \end{tabular}
    \label{tab:perturbations_uncertainties_and_methods}
    \vspace{-1.5em}
\end{table}

\subsubsection{Uncertainty-Aware Action Selection}
For collision avoidance, the uncertainty estimates are used for a \ac{POC} estimation using \eqref{eq:prob_col_combined} with the threshold constants for \ac{MC-Dropout} $\lambda_{\mathrm{ep}}=0.03$, $\lambda_\mathrm{f} = 0.3$, $\lambda_{\mathrm{prox}}=0.9$, $\beta_1=0.5$ and deep ensemble $\lambda_{\mathrm{ep}}=0.08$, $\lambda_\mathrm{f} = 0.0033$, $\lambda_{\mathrm{prox}}=0.9$, $\beta_1=0.5$.
The parameters are identified by analyzing uncertainty-perturbation correlations, e.g. Fig.~\ref{fig:epistemic_aleatoric_uncertainty} and Fig~\ref{fig:perturbations_and_uncertainties}.
This ensures that the safe action selection only activates in steps or scenarios with a high risk of collision. 
The results show that the \ac{MC-Dropout} is more suitable for safe action selection, as aggregated in Table~\ref{tab:prevented_collisions} with relative percentage of prevented collisions compared to no safe action selection. 

\begin{table}[!htb]
\centering
\vspace{-0.5em}
\caption{Percentage of prevented collisions with safe action selection.}
\vspace{-0.5em}
\setlength{\tabcolsep}{3.5pt}
\begin{tabular}{|c|c|c|c|c|}
\hline
Approach    &   Obs. Noise                  &   Action Noise                & Velocity              &   More Humans \\ \hline \hline
MC-dropout  &   $\bm{100} \pm \bm{0}$     & $\bm{6} \pm \bm{5}$             & $\bm{56} \pm \bm{2}$  &   $\bm{81} \pm \bm{8}$ \\
Ensemble    & ${79} \pm {11}$        & ${0} \pm {0}$            & ${16} \pm {1}$  & ${59} \pm {13}$ \\                                             
 \hline

\end{tabular}
\label{tab:prevented_collisions}
\vspace{-1.5em}
\end{table}

\subsection{Discussion}
In social navigation, human safety is influenced by physical aspects (e.g. collisions) and psychological aspects (e.g. discomfort). 
Our reward formulation accounts for both, and thus maximizing the reward also improves the safety in both aspects. 
Using dropout and \ac{ODV} in \ac{PPO} leads to better policy quality in terms of training convergence, evaluation, and generalization performance. 
With a higher success rate and a lower collision rate, the \ac{ODV-PPO} with dropout approach is also safer by design. 
In addition, \ac{ODV} improves the social navigation twofold. 
First, it enables the disentangling of the uncertainty, and second, it enables observation-specific exploration-exploitation trade-off. 
In particular, the latter improves social navigation since the policy can explore more in complex human-robot interaction situations and less in simple navigation situations. 
The circle interaction scenario with staged interaction has the least stochasticity in position sampling but leads to good generalization, faster training convergence, and lower training variance across all investigated algorithms. 
This opens the research question of whether the staged interaction in training can support generalization while leading to fast convergence.
Nevertheless, if the scenario distribution differs from the training distribution, the performance decreases, and the policy predicts overconfident actions.
However, a predictive uncertainty does not necessarily correlate with high uncertainty in decision-making as visualized in Fig.~\ref{fig:stepwise_uncertainty}, but can reveal risky scenarios.
This further confirms the need for disentangling uncertainty estimates in policy distribution and not only to detect novel scenarios or predictive uncertainty. 
The comparison of the uncertainty methods shows that the \ac{MC-Dropout} approach can better distinguish between aleatoric and epistemic uncertainty and has additionally higher sensitivity to noise than the ensemble approach.
However, the experiments revealed that a properly selected dropout rate in inference is crucial for this behavior. 
The aleatoric uncertainty is sensitive to noise, showing that it properly associates the uncertainty type with the source of uncertainty.
However, since all perturbed scenarios are partially \ac{OOD} scenarios and the policy was only trained in a very simple scenario, the epistemic uncertainty is generally high and correlates with aleatoric uncertainty. 
These correlation findings for disentangling uncertainty in \ac{DRL} are similar to recent work on classification \cite{MucsanyiBenchmarkingUncertaintyDisentanglement2024}, where aleatoric and epistemic uncertainty show correlation, but tailored methods can be used for specific applications.
Furthermore, since an accurate aleatoric uncertainty estimate requires low epistemic uncertainty, the aleatoric can only be estimated reliably in low epistemic areas \cite{AndreyMalininPredictiveUncertaintyEstimationviaPr2018}. 
In addition, the results show that the perturbation sources have varying impacts on decision-making, whereas they have an equal impact on predictive uncertainty.   
The policy is more uncertain about the heading throughout all perturbations, which is crucial for safe action selection and future design of \ac{DRL} policies. 
More humans in the environment cause the highest uncertainty in decision-making, but collisions are most difficult to prevent in the case of action noise. 
For safe action selection, the results show that it is possible to parametrize a threshold function for \ac{POC} calculation to detect scenarios of high risk to prevent them. 
However, one downside of this approach is the manual parametrization, which requires a perturbation analysis in advance. 
Furthermore, if the uncertainty estimation method has low sensitivity for perturbation, it is more challenging for the \ac{POC} as in the ensemble approach.

\section{Conclusions} 
This paper incorporated and disentangled epistemic, aleatoric, and predictive uncertainty estimation in a \ac{DRL} policy with a safe action selection for perturbation robustness in social navigation. 
We integrated \acf{ODV} and dropout into the \ac{PPO} algorithm with adaptions in the action network and loss function.
This, and staged interaction in training scenarios, lead to a better generalization with faster convergence and enables disentangling the aleatoric and epistemic uncertainty in decision-making.
The results show that the \ac{MC-Dropout}-based approach is superior for uncertainty estimation and, in combination with the proposed safe action selection, can avoid more collisions than the ensemble approach. 
In addition, the predictive uncertainty in perturbed environments can be high, although the robot is sure about selecting the action. 
Consequently, \ac{MC-Dropout} and \ac{ODV} should be used to estimate the policy distribution uncertainty to detect when and why the robot is uncertain. 
Future works will develop sample-free methods.

\addtolength{\textheight}{-0cm}   

\bibliographystyle{IEEEtran}
\bibliography{indices/references}

\end{document}